\documentclass[conference]{IEEEtran}
\IEEEoverridecommandlockouts
\usepackage{cite}
\usepackage{amsmath,amssymb,amsfonts}
\usepackage{algorithmic}
\usepackage{graphicx}
\usepackage{subfig}
\usepackage{stfloats}
\usepackage{adjustbox}
\usepackage{textcomp}
\usepackage[hidelinks,pagebackref]{hyperref}
\hypersetup{
    colorlinks = true,
    linkcolor=blue,       % Internal cross-references, TOC, etc.
    citecolor=blue,        % Citation links
    urlcolor=blue          % Web links
}
\usepackage{xcolor}
\usepackage{xspace}
\begin{document}
\title{Demonstration-Guided Humanoid Stand-Up on an Emulated Deformable Surface}
%
% \author{\IEEEauthorblockN{Aniruddh Kushwah, Vyankatesh Ashtekar, and Ashish Dutta}
% \IEEEauthorblockA{\textit{Department of Mechanical Engineering} \\
% \textit{Indian Institute of Technology Kanpur}\\
% Kanpur, Uttar Pradesh, India \\
% \{aniruddhk21, vyankatesh20, adutta\}@iitk.ac.in}
% }
\author{%
\IEEEauthorblockN{Aniruddh Kushwah, Vyankatesh Ashtekar\thanks{Corresponding author: Vyankatesh Ashtekar vyankatesh20@iitk.ac.in}, and Ashish Dutta}
\IEEEauthorblockA{\textit{Department of Mechanical Engineering} \\
\textit{Indian Institute of Technology Kanpur}\\
Kanpur, Uttar Pradesh, India \\
\{aniruddhk21, vyankatesh20, adutta\}@iitk.ac.in}
}
\maketitle
\begin{abstract}
This paper presents a reference-guided reinforcement learning framework to generate stand-up motion for a 29-DOF Unitree G1 humanoid on deformable soft ground, using a human demonstration recorded on hard ground. The terrain compliance is modelled using \texttt{solref} and \texttt{solimp} parameters from MuJoCo's rigid body soft-contact model. 
The rewards consists of (i) reference motion tracking through residual joint-position control and (ii) explicit recovery objectives such as pelvis height, torso uprightness, and the final posture.
First, the policy is trained with the specified rewards considering hard ground. Next, the terrain stiffness is lowered by updating \texttt{solref} and the nominal surface penetration zone is expanded using \texttt{solimp}. Subsequent training enables the policy to adapt to the delayed support force generation due to significant surface penetration during contact-intensive phases while preserving the original demonstration pattern.
The learned policy successfully completes the fallen-to-standing task in simulation, reaching the targeted pelvis height and uprightness, with a maximum contact penetration of approximately 40~mm during the process.
The proposed method is demonstrated on two stand-up sequences and successfully achieves the final recovery objective on both hard and soft ground.
Ablation studies show that reference tracking alone is insufficient for successful stand-up, and that explicit recovery rewards are essential.
\end{abstract}
\section{Introduction}
% Introduce: stand-up + soft ground
Long-term autonomy of humanoid robots in industrial or outdoor environments depends on reliable autonomous stand-up capability.
Compared to locomotion, standing up from a lying posture is a contact-rich non-cyclic task.
Extending stand-up motion synthesis to deformable surfaces such as foam slab or thick grass is particularly challenging because large deformations are necessary on soft terrain to generate the required contact forces, significantly altering the robot's posture and balance.
% One line: what does this paper do
We introduce a reinforcement learning~(RL) framework that leverages stand-up demonstrations, which are easier to collect on hard surfaces, and adapts them for execution on deformable soft terrain.

% Literature review starts here:
The motion synthesis methods for getting up from a lying pose can be broadly organised into four categories: (a)~preplanned action sequences~\cite{Fujiwara2003, Hirukawa2005,Kanehiro2007,Araki2018,Stelter2021} (b)~optimisation-based discovery of motion~\cite{Tassa2012_IROS_TO,Mordatch2012_CIO,Howell2022}, (c)~RL-based discovery of motion~\cite{Gaspard2025,Chen2025,Huang2025_RSS} and (d)~reference motion guided methods, such as imitation learning~\cite{Peng2021, Tang2024,Peng2018}.

% (a) predefined
Earlier research on humanoids getting-up from lying position relied primarily on predefined motion sequences, contact-state-based control and key-pose interpolation. Fujiwara~et~al.~\cite{Fujiwara2003} developed an early human-sized humanoid capable of mitigating fall impact and subsequently standing up. Hirukawa~et~al.~\cite{Hirukawa2005} demonstrated a humanoid platform capable of walking, lying down, and getting up autonomously. Kanehiro~et~al.~\cite{Kanehiro2007} proposed a getting-up motion-planning method based on the Mahalanobis distance. Araki~et~al.~\cite{Araki2018} planned the rising motion by selecting actions according to the robot’s current contact configuration with the ground,
Araki~et~al.~\cite{Araki2018} generated standing-up motions according to the robot’s body-ground contact state. Stelter~et~al.~\cite{Stelter2021} developed fast and reliable closed-loop stand-up motions using parametrised trajectory interpolation.

%(b) Discovery of stand up sequency via optimisation or RL
Tassa~et~al.~\cite{Tassa2012_IROS_TO} formulated stand-up motion synthesis as a non-linear model predictive control~(MPC) problem solved using iterative Linear Quadratic Gaussian~(iLQG) trajectory optimisation while accounting for contact dynamics via MuJoCo~\cite{todorov2012mujoco}. 
The state cost comprised penalties on the horizontal projection of distances to promote standing-up behaviour, e.g., Centre of Mass (COM)-centre of support, torso-COM, torso-desired height of stance. Robot's horizontal COM velocity was penalised too. Stiff contacts improved realism, while softening of contacts smoothened the derivatives of contact dynamics, aiding quicker optimisation convergence. Recently released MuJoCo-MPC~\cite{Howell2022} provides a wider framework for implementing multiple MPC algorithms with MuJoCo.
Mordatch~et~al.~\cite{Mordatch2012_CIO} proposed contact-invariant optimisation (CIO), in which contact activation variables are optimised together with the motion trajectory, allowing the optimiser to automatically determine contact timings and transitions for complex whole-body behaviours. 
%below was flagged as plag
% Heuristic sub-goals and hint cost terms are required in the formulation for achieving useful movements such as getting up. 
The formulation still requires manually designed intermediate objectives and auxiliary costs to guide the robot toward meaningful behaviours, including standing up. The study focused on organising the exploration space of variables, leaving the full dynamic implementation of motion to the future.

RL has increasingly been applied to the discovery of contact-rich motions because sampling-based methods can operate in the presence of discontinuous contact transitions.
Jeong and Lee~\cite{Jeong2016} exploited the bilateral symmetry of humanoid robots by sharing Q-learning updates between mirror-symmetric states and actions, thereby reducing the amount of exploration required and improving learning efficiency.  
%below was flagged as plag
% Gaspard~et~al.~\cite{Gaspard2025} introduced FRASA, an end-to-end RL framework for humanoid fall recovery and standing-up. 
Gaspard~et~al.~\cite{Gaspard2025} proposed FRASA, a unified RL approach that learns the complete transition from a fallen posture to standing. Huang~et~al.~\cite{Huang2025_RSS} proposed HoST, which learns standing-up behaviours across diverse initial postures using curriculum learning and specialised policy-training mechanisms. 
%below was flagged as plag
% Chen~et~al..~\cite{Chen2025} developed HiFAR, a multi-stage curriculum-learning framework for high-dynamic humanoid fall recovery. 
Chen~et~al.~\cite{Chen2025} presented HiFAR, which uses a staged curriculum to train highly dynamic recovery motions for humanoid robots. He~et~al.~\cite{He2025} also demonstrated RL-based humanoid recovery using staged training, exploration assistance, initial-pose randomisation, and reward shaping.
Collectively, these approaches employ curriculum learning, assistance forces, staged policy refinement, initial-state randomisation, multi-critic learning, and carefully designed reward functions to obtain recovery behaviours from prone, supine, and intermediate configurations. Some methods first discover a feasible recovery motion and subsequently refine it to satisfy smoothness, torque, and deployment constraints, while others train a single end-to-end policy or generalise recovery across multiple initial states or robot morphologies. Despite these advances, the cited recovery methods do not explicitly study the transfer of a demonstrated stand-up motion from rigid ground to a substantially compliant support surface. When terrain variation is considered, compliance is typically included as one of the many dynamics randomisation parameters being tweak lightly, rather than treated as the primary contact-control challenge. Consequently, these works provide limited insight into how large surface deformation, delayed support-force generation, altered contact geometry, and energy dissipation affect the execution of a reference stand-up trajectory. Motion-imitation methods likewise generally assume rigid support and do not directly adapt demonstrated contact sequences to markedly reduced ground stiffness and damping. To the best of our knowledge, none of these methods~\cite{Gaspard2025,Chen2025,Huang2025_RSS,spraggett2025learningmorphologieszeroshotrecovery,poddar2026embeddingclassicalbalancecontrol,lu2026unifiedwalkingrunningrecovery} specifically investigate RL-based stand-up behaviour under substantially compliant contact conditions.

Motion-capture-guided imitation learning provides an alternative by introducing a human-motion prior into policy learning. 
Peng~et~al.~\cite{Peng2018} introduced DeepMimic, which uses RL to track reference motions through objectives defined over joint poses, joint velocities, end-effector positions, and root motion.
%below was flagged as plag
% Peng~et~al.~\cite{Peng2021} later proposed Adversarial Motion Priors, in which a learned discriminator provides a motion-quality reward that encourages the policy to remain within the distribution of demonstrated human motions.
Peng~et~al.~\cite{Peng2021} subsequently introduced Adversarial Motion Priors, where a discriminator evaluates motion realism and rewards behaviours that resemble the demonstrated human-motion distribution.
Tang~et~al.~\cite{Tang2024} extended motion-imitation and adversarial-prior methods to humanoid whole-body control, while 
He~et~al.~\cite{He2025} incorporated motion references into learned recovery behaviours. Luo~et~al.~\cite{Luo2026} learned motion priors from diverse motion-capture data using dense supervision and released a large motion dataset, although its demonstrations are limited to rigid ground.
% ~\cite{Luo2026} learned human motion priors from a large and diverse motion-capture corpus using dense supervision, reducing reliance on manually engineered rewards. They also introduced a rich motion dataset, although its demonstrations are limited to rigid-ground settings.
%below was flagged as plag
% In these approaches, human motion is retargeted to the robot, and reinforcement learning adapts the reference trajectory to the robot’s morphology, actuation limits, contact dynamics, and physical environment.
These methods first map human demonstrations to the robot and then use RL to modify the motion according to its body structure, actuator constraints, contact behaviour, and surroundings.
Explicit trajectory-tracking approaches provide detailed temporal guidance, whereas adversarial motion-prior methods encourage human-like behaviour without requiring strict frame-by-frame reproduction. These techniques have produced coordinated and natural humanoid locomotion, transitions, and whole-body skills. However, most existing motion-imitation studies are evaluated on rigid ground and focus primarily on locomotion, teleoperation, dancing, or loco-manipulation rather than recovery from a fully fallen configuration.

Compliant ground introduces additional challenges because deformation under load alters the effective contact geometry, centre of pressure, support region, contact wrench, energy dissipation, and balance response of the robot.
%below was flagged as plag
% Mesesan~et~al.~\cite{Mesesan2019} investigated dynamic humanoid walking on compliant and uneven terrain using Divergent Component of Motion (DCM) and passivity-based whole-body control
Mesesan~et~al.~\cite{Mesesan2019} studied dynamic locomotion over soft and irregular surfaces by combining Divergent Component of Motion (DCM) planning with passivity-based whole-body control, treating softness as an unmodeled disturbance handled through stance-foot damping.
Singh~et~al.~\cite{Singh2024_compliant} applied RL to humanoid locomotion over compliant terrain. The randomised~\texttt {solref} parameter to simulate hard as well as soft terrain.
% , while Lynch~et~al.~\cite{Lynch2025} examined dynamic behaviours such as hopping and balancing on deformable surfaces.\tbd{use this Lynch2025 only if you need some citation on foam penetration etc. no need to cite directly.}
Existing learned getting-up systems, including HUMANUP~\cite{He2025} and related recovery frameworks, have also demonstrated recovery on surfaces such as grass, snow, mattresses, or irregular terrain. 
It remains to be investigated how a recorded stand-up demonstration should be adapted for conditions with significant terrain compliance.

% Main contribution
The main contribution of this work is a two-stage RL training method that adapts human stand-up demonstrations recorded on hard ground to synthesize stand-up motions for a 29-DOF Unitree G1 humanoid on deformable soft terrain.
First, a policy is trained to track a retargeted human demonstration using motion-tracking rewards together with explicit task-level objectives for pelvis height, torso uprightness, and the final standing posture. 
The policy interacts with a MuJoCo simulation employing a rigid body soft contact model with the default solver reference (\texttt{solref}) and impedance (\texttt{solimp}) parameters representing a hard ground. 
Next, the training is continued on softened terrain, achieved by appropriately updating the \texttt{solref} and \texttt{solimp}. Also, the learning hyperparameters are retuned systematically, and a reset noise curriculum is introduced for robustness.
Ablation studies are conducted to ascertain the necessity of the proposed rewards. The proposed method is validated on two different stand-up sequences. The resulting policy works on hard as well as soft ground.
\section{Method}
\label{sec:Method}
The proposed method learns a human-like fallen-to-standing motion for the 29-DOF Unitree G1 humanoid using reference-guided RL. Two stand-up trajectories from the BONES-SEED dataset~\cite{Luo2026}, already retargeted to the Unitree G1 using the kinematic retargeting-based method General Motion Retargeting (GMR)~\cite{araujo2025retargetingmattersgeneralmotion} and PyRoki~\cite{11246651} toolkit, are used as the motion reference. A separate policy is trained for each trajectory and then fine-tuned on a compliant surface modelled using MuJoCo soft-contact parameters. Proximal Policy Optimisation (PPO)~\cite{schulman2017proximalpolicyoptimizationalgorithms} is used to learn residual joint-position commands, while joint-level Proportional-Derivative (PD) controllers execute the commanded motion.
\subsection{Reinforcement Learning Formulation}
%below was flagged as plag
% The stand-up task is formulated as a finite-horizon Markov decision process (MDP).
We model the stand-up problem as a Markov decision process (MDP) with a fixed episode duration.
The policy operates at~$60~\mathrm{Hz}$ (frame-skip 8 over a 2 ms integrator step), while the reference trajectory is sampled at~$120~\mathrm{Hz}$. At every control step, the reference frame is selected according to the elapsed simulation time.
%below was flagged as plag
% \ani{rephrase later:}The policy outputs a normalized action vector~$\mathbf{a}_t \in [-1,1]^{29}$ which represents a residual correction to the reference joint configuration. 
The actor network returns a 29-dimensional bounded control signal, with each element specifying a residual adjustment to the corresponding reference joint position. The commanded joint positions are:
\begin{align}
    \mathbf{q}^{\mathrm{cmd}}_t = \mathbf{q}^{\mathrm{ref}}_t + s_a\mathbf{a}_t,
\end{align}
where~$\mathbf{q}_t$ denotes the joint angle vector,~$\mathbf{a}_t$ is the action vector, and~$s_a=0.25$ is a re-scaler. The resulting commands are clipped to the actuator limits and tracked using joint-level PD control:
\begin{align}
    \tau_{j,t} = K_{p,j} \left( q^{\mathrm{cmd}}_{j,t} - q_{j,t} \right) - K_{d,j}\dot{q}_{j,t}.
\end{align}
The hip joints use~$K_p=150$~N$\cdot$m/rad and~$K_d=4$~N$\cdot$m$\cdot$s/rad, the knees use~$K_p=200$ and~$K_d=6$, the ankles use~$K_p=200$ and~$K_d=2$, and the waist and upper-body joints use~$K_p=100$ and~$K_d=4$.

The policy observation contains the simulated state, corresponding reference state, reference-tracking errors, deviation from the final standing posture, motion phase, and previous action:
\begin{align}
    \mathbf{o}_t = \left[ \mathbf{q}^{\mathrm{sim}}_{t,2:}, \dot{\mathbf{q}}^{\mathrm{sim}}_t, \mathbf{q}^{\mathrm{ref}}_{t,2:}, \dot{\mathbf{q}}^{\mathrm{ref}}_t, \mathbf{e}^{q}_t, \mathbf{e}^{\dot q}_t, \mathbf{e}^{\mathrm{stand}}_t, \phi_t, \mathbf{a}_{t-1} \right].
\end{align}
The error terms are defined as~$\mathbf{e}^{q}_t = \mathbf{q}^{\mathrm{sim}}_t - \mathbf{q}^{\mathrm{ref}}_t, \mathbf{e}^{\dot q}_t = \dot{\mathbf{q}}^{\mathrm{sim}}_t - \dot{\mathbf{q}}^{\mathrm{ref}}_t,$ and~$\mathbf{e}^{\mathrm{stand}}_t = \mathbf{q}^{\mathrm{sim}}_t - \mathbf{q}^{\mathrm{stand}}$. 
The normalised motion phase~$\phi_t \in [0, 1]$ indicates progress through the reference trajectory, from the first frame ($0$) to the final frame ($1$), providing temporal context to the policy.
\begin{align}
\phi_t = \frac{i_t}{N-1},
\end{align}
where~$i_t$ is the current reference-frame index and~$N$ is the trajectory length.

The policy and value networks each contain three hidden layers of 512, 256, and 128 units.
Training uses 20 parallel environments, a discount factor of~$0.995$, Generalised Advantage Estimation (GAE) coefficient~$0.95$, clipping threshold~$0.2$, batch size~$2048$, and five optimisation epochs.
\subsection{Reference-Guided Training and Adaptation to the Compliant Terrain}
The motion-capture data needs to be preconditioned before using it as the reference trajectory. The stand-up motion is obtained from the publicly available BONES-SEED dataset~\cite{Luo2026}. The pelvis is treated as the floating base, and its root translations need to be rescaled appropriately. The global height of the motion is then manually corrected, and the reference root position and orientation are aligned with the fallen key frame of the MuJoCo model. Reference joint velocities are estimated by finite differencing the motion capture trajectory data.

The reward combines trajectory tracking, final-standing objectives, and regularisation terms. The complete reward is
\begin{equation}
\begin{split}
r_t &= w^{\mathrm{track}}_t r^{\mathrm{track}}_t
     + w_t r^{\mathrm{final}}_t \\
    &\quad + 0.2
     - 0.15p^{\mathrm{slip}}_t
     - 0.002p^{\mathrm{ctrl}}_t
     - 0.005p^{\mathrm{smooth}}_t .
\end{split}
\end{equation}
Table~\ref{tab:reward_terms} lists the definitions and coefficients of the reward components. Further details are available in the project \href{https://github.com/andireposit/Stand-Up-Motion-on-Compliant-Surface-for-Humanoid}{repository}.
\begin{table}[t]
\centering
\caption{Reward terms used for reference tracking and final stand-up completion.}
\label{tab:reward_terms}
\resizebox{\columnwidth}{!}{
\begin{tabular}{lll}
\hline
\textbf{Term} & \textbf{Definition} & \textbf{Weight} \\
\hline
\multicolumn{3}{l}{\textit{Reference-Tracking Rewards}} \\
Joint pose tracking
&
$\displaystyle r^{\mathrm{pose}}_t = \exp \left[-4 \operatorname{mean} \left( (\mathbf{q}^{\mathrm{sim}}_t - \mathbf{q}^{\mathrm{ref}}_t)^2 \right)\right]$
&
$2.0$
\\
Joint velocity tracking
&
$\displaystyle r^{\mathrm{vel}}_t = \exp \left[-0.03 \operatorname{mean} \left( (\dot{\mathbf{q}}^{\mathrm{sim}}_t - \dot{\mathbf{q}}^{\mathrm{ref}}_t)^2 \right)\right]$
&
$0.7$
\\
Root planar tracking
&
$\displaystyle r^{xy}_t = \exp \left[-2 \operatorname{mean} \left( (\mathbf{p}^{xy}_t - \mathbf{p}^{xy,\mathrm{ref}}_t)^2 \right)\right]$
&
$0.2$
\\
Pelvis height tracking
&
$\displaystyle r^{h}_t = \exp \left[-10(h_t - h^{\mathrm{ref}}_t)^2\right]$
&
$1.0$
\\
Root orientation tracking
&
$\displaystyle r^{\mathrm{ori}}_t = \exp(-\theta_t^2)$
&
$0.5$
\\
\hline
\multicolumn{3}{l}{\textit{Standing rewards, phase $> 0.65$}} \\
Standing-pose tracking
&
$\displaystyle r^{\mathrm{standpose}}_t = \exp \left[-5 \operatorname{mean} \left( (\mathbf{q}^{\mathrm{sim}}_t - \mathbf{q}^{\mathrm{stand}})^2 \right)\right]$
&
$2.0$
\\
Uprightness
&
$\displaystyle r^{\mathrm{up}}_t = \operatorname{clip} \left( \frac{\mathbf{z}^{p}_t \cdot \mathbf{z}^{w}+1}{2}, 0, 1 \right)$
&
$2.0$
\\
Standing height
&
$\displaystyle r^{\mathrm{standheight}}_t = \exp \left[-20(h_t - h^{\mathrm{stand}})^2\right]$
&
$1.5$
\\
\hline
\multicolumn{3}{l}{\textit{Regularization Penalties}} \\
Foot-slip penalty
&
$\displaystyle p^{\mathrm{slip}}_t = \sum_{f\in\mathcal{F}_c} |\mathbf{v}^{xy}_{f,t}|^2$
&
$-0.15$
\\
Action penalty
&
$\displaystyle p^{\mathrm{ctrl}}_t = \operatorname{mean}(\mathbf{a}_t^2)$
&
$-0.002$
\\
Action-smoothness penalty
&
$\displaystyle p^{\mathrm{smooth}}_t = \operatorname{mean} \left[ (\mathbf{a}_t - \mathbf{a}_{t-1})^2 \right]$
&
$-0.005$
\\
\hline
\end{tabular}}
\end{table}
The tracking reward is
\begin{align}
\label{tracking-only-eq}
r^{\mathrm{track}}_t = 2r^{\mathrm{pose}}_t + 0.7r^{\mathrm{vel}}_t + 0.2r^{xy}_t + r^{h}_t + 0.5r^{\mathrm{ori}}_t,
\end{align}
and the final-standing reward is
\begin{align}
r^{\mathrm{final}}_t = 2r^{\mathrm{standpose}}_t + 2r^{\mathrm{up}}_t + 1.5r^{\mathrm{standheight}}_t.
\end{align}
A phase-dependent weight gradually shifts the objective toward final standing:
\begin{align}
w_t = \operatorname{clip} \left( \frac{\phi_t-0.65}{0.35}, 0, 1 \right), \qquad w^{\mathrm{track}}_t = 1 - 0.5w_t.    
\end{align}

The policy is initially trained for approximately~$70$ million simulation steps and is subsequently fine-tuned on compliant ground for another~$20$ million steps. During fine-tuning, the learning rate is reduced from~$3\times10^{-4}$ to~$1\times10^{-4}$, and the entropy coefficient is reduced from~$0.003$ to~$0.001$. Observation and reward normalisation statistics from the pre-trained model are retained.

A reset-noise curriculum is used during compliant-ground fine-tuning. Gaussian noise is applied to the initial joint positions and generalised velocities. The noise levels increase linearly with training progress~$p$:
\begin{align}
\sigma_q(p) = 0.005 + p(0.030 - 0.005), \\
\sigma_{\dot q}(p) = 0.005 + p(0.100 - 0.005).
\end{align}
The initial reference frame is sampled from the first six frames of the trajectory. Root-position noise, yaw noise, external pushes, gravity randomisation, and friction randomisation are disabled.

The compliant floor is modelled by repurposing MuJoCo's phenomenological soft contact formulation rather than an explicit constitutive law. In MuJoCo, rigid contact is softened by relaxing the strict complementarity between contact force and interpenetration. The resulting normal contact dynamics can be interpreted approximately as follows~\cite{MuJoCo_SolverParameters}:
\begin{align}
&a_c \approx \gamma(r_c) a_{\mathrm{ref}} + \left[ 1 - \gamma(r_c) \right] a_{\mathrm{free}}, \text{ with,}\\
&a_{\mathrm{ref}} = -\kappa \, r_c - c\,\dot{r}_c
\end{align}
where,~$a_c$ denotes the normal acceleration associated with the contact constraint,~$a_{\mathrm{free}}$ is the normal acceleration that would occur without contact,~$r_c$ is the contact penetration depth~(or more appropriately, the contact residual), and~$\dot{r}_c$ is its rate of change. The function~$\gamma(r_c) \in (0, 1)$ determines the solver impedance, i.e., the extent to which the contact constraint is enforced as penetration increases. The parameters~$\kappa$ and~$c$ represent the reference stiffness and damping, with units~$s^{-2}$ and~$s^{-1}$, respectively.
For~$\gamma(r_c)$ close to one, the normal response is dominated by the spring--damper term. 
Smaller values produce a softer response in which the acceleration remains closer to the unconstrained dynamics.
The default contact parameters are:
\begin{align}
\texttt{solref}=(0.02, 1), \,
\texttt{solimp}=(0.9, 0.95, 0.001).
\end{align}
The~$\texttt{solimp}$ consists of two more parameters, but they were left unchanged. The updated parameters for simulating compliant terrain are as follows:
\begin{align}
\texttt{solref}=(0.1, 1), \,
\texttt{solimp}=(0.0, 0.95, 0.02).
\end{align}
A critically damped response with a~$0.1~\mathrm{s}$ time constant, and solver impedance increasing from~$0$ to~$0.95$ over~$0.02~\mathrm{m}$ depth is chosen. 
The \texttt{width}($=0.02$~m) parameter does not impose a hard penetration limit; rather, it specifies the penetration depth over which the impedance increases from its initial to its maximum specified value, remaining approximately constant thereafter.
Thus, the model approximates compliant ground through gradual constraint enforcement while retaining a planar floor geometry. The chosen width parameter is realistic, corresponding to a foam slab, say 5~cm thick and representative human body loading~\cite{Takashi2018}.

Each episode includes the reference-motion duration and an additional~$2~\mathrm{s}$ standing interval. It terminates if any of the elements in the state vector becomes non-finite (\texttt{Nan} or \texttt{Inf}), indicating numerical instability, or if the pelvis height~$h_t \notin \left[-0.1, 1.8\right]$~m.
Model performance is assessed using pelvis height, uprightness, motion completion, contact penetration, and stand-up snapshots.
\section{Results and Discussion}
\begin{figure}[t]
\centering
\includegraphics[width=\columnwidth]{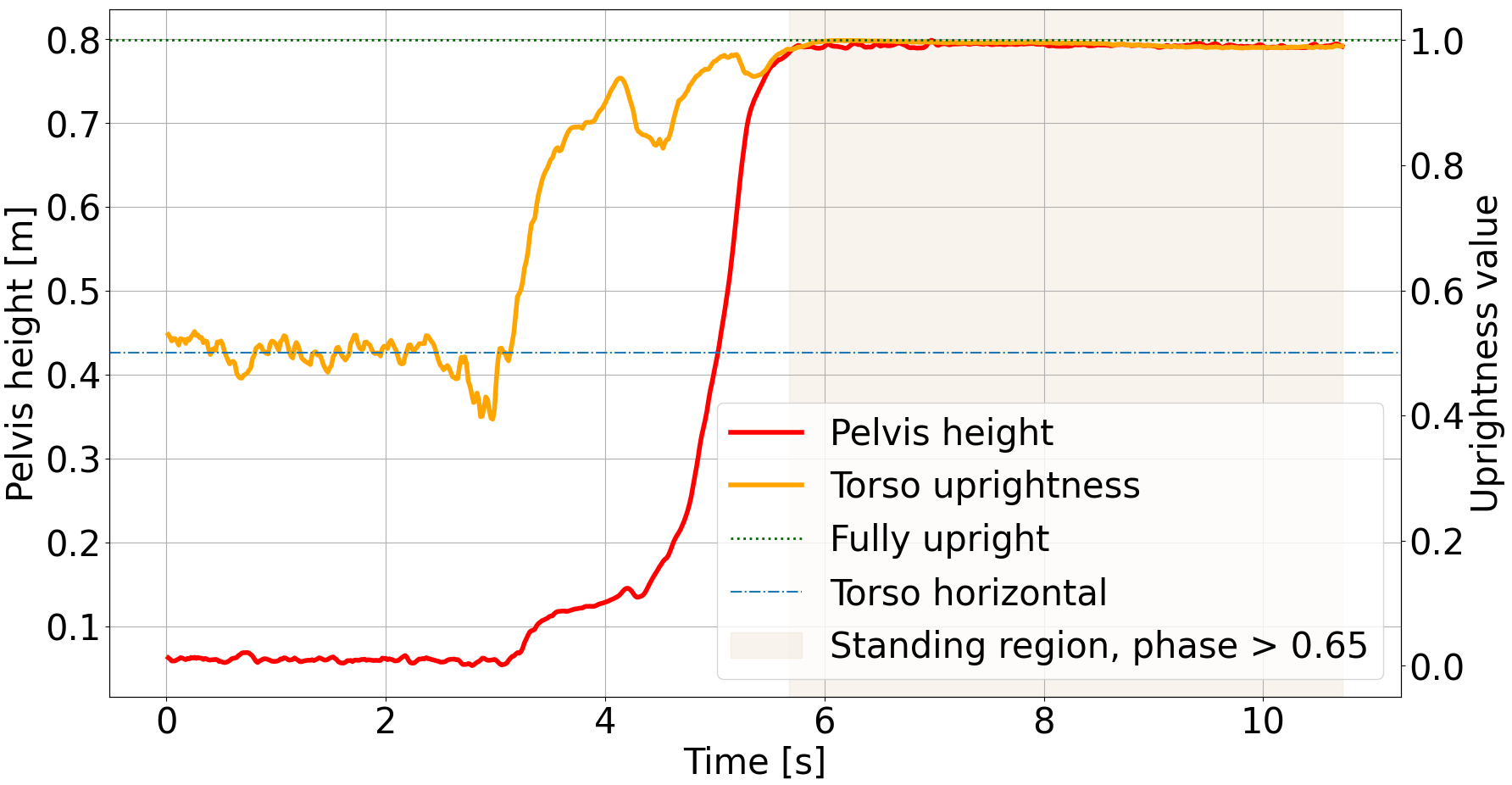}
\caption{Pelvis height and torso uprightness for one representative policy.}
\label{fig:height_uprightness}
\end{figure}

\begin{figure*}[t]
\centering
\includegraphics[width=\textwidth]{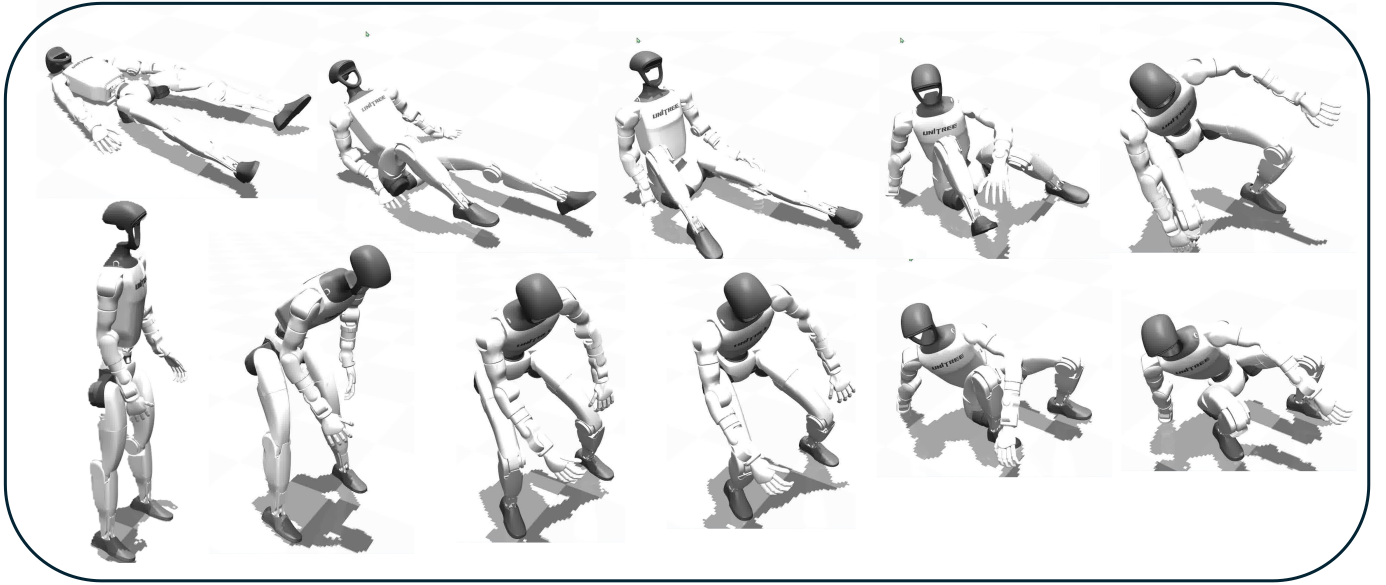}
\caption{Sequential snapshots of the stand-up policy in MuJoCo simulation environment.}
\label{fig:stand_sequence}
\end{figure*}

\begin{figure}[thb]
\centering
\includegraphics[width=\columnwidth]{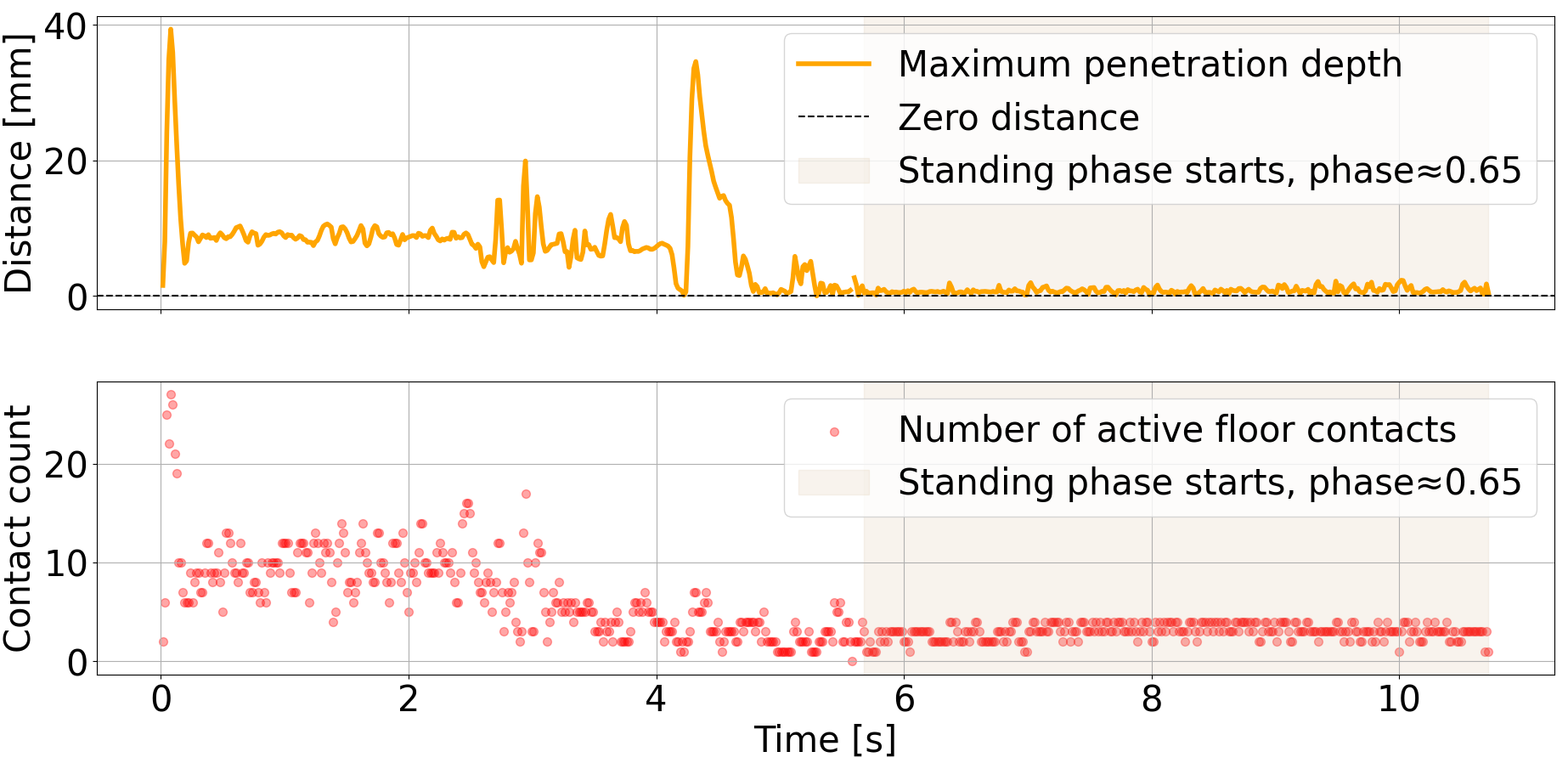}
\caption{Active floor contacts for one representative policy: penetration depth and active-contact count. The maximum penetration
was approximately~$40$~mm.}
\label{fig:contact_penetration}
\end{figure}

The proposed method was applied to two retargetted human demonstrations varying in initial position and stand-up sequence as shown in the supplementary
\href{https://youtu.be/c04fnMCDdd8}{video}.
Both the policies successfully completed their respective fallen-to-standing motion on the hard as well as soft flat surfaces.
Figures~\ref{fig:height_uprightness}, \ref{fig:stand_sequence}, \ref{fig:contact_penetration} show representative results from a deterministic policy inference. 
One can observe that the demonstrated movement sequence remains largely unaltered. Small deviations in joint-space appear during contact-intensive phases. These deviations are a result of policy accommodating terrain softness into the motion while maintaining balance.

Task completion was evaluated quantitatively using pelvis height and torso uprightness. As shown in Fig.~\ref{fig:height_uprightness}, the pelvis height increased progressively from the fallen configuration and rose sharply during the final lower-body extension phase. The achieved pelvis height was~$0.792~\mathrm{m}$, compared with the target value of~$0.794~\mathrm{m}$.
The final uprightness value was recorded as~$0.991$, indicating that the body-fixed vertical axis of the torso body was closely aligned to the world vertical direction. The combined height and uprightness results confirm that the policy did not merely lift the pelvis but achieved the expected posture.

The compliant-contact behaviour was evaluated using contact measurements recorded during policy inference~(see e.g., Fig.~\ref{fig:contact_penetration}) and a separate drop test shown in the supplementary video.
This behaviour results from the low initial impedance of the contact model, which allows penetration before the constraint response becomes stronger.
The active-contact analysis in Fig.~\ref{fig:contact_penetration} provides a quantitative view of this behaviour during the learned stand-up motion. For every active contact involving the floor, the world-frame contact-point penetration was recorded. The largest recorded value during policy inference was~$\delta_{\max} = 39.4~\mathrm{mm}.$
The number of active contacts also changed considerably throughout the motion. The full body contacted the ground at multiple locations in the initial fallen phase, while the final stage was marked by only the feet in contact. Penetration depth increases as the humanoid tries to stand up~(see e.g., Fig.~\ref{fig:contact_penetration}), making the movement difficult as compared to the hard ground.
% It increased when the humanoid loaded the surface to raise or reposition the body and decreased when contacts were unloaded or removed. The signed contact distance is the direct MuJoCo measure of contact-constraint violation, while the contact-point height provides a geometric representation relative to the floor surface. Together, these measurements confirm that the stand-up motion was executed under substantial compliant contact rather than near-rigid support.

The reset-noise curriculum also contributed to the robustness of the learned behaviour. The policy continued to complete the motion as the initial joint-position noise increased to~$0.030~\mathrm{rad}$ and the velocity-noise magnitude increased to~$0.100$. Small variations in the initial state can change the first contact locations, penetration depth, and timing of force generation, particularly on compliant ground. Training under these perturbations, therefore, reduced the policy's dependence on one exact fallen configuration. However, the demonstrated robustness remains local to the applied position and velocity noise ranges, since yaw perturbations, horizontal root displacement, external pushes, gravity variation, and friction randomisation were not included.

For ablation study, we compared tracking-only and complete-reward policies under identical simulation and PPO settings. 
% The tracking-only policy used only the reference-tracking reward defined in~\eqref{tracking-only-eq},
The first policy used only the reference-tracking component,
\begin{align}
r^{\mathrm{track}}_t = 2r^{\mathrm{pose}}_t + 0.7r^{\mathrm{vel}}_t + 0.2r^{xy}_t + r^{h}_t + 0.5r^{\mathrm{ori}}_t,    
\end{align}
% whereas the working policy used the complete phase-dependent reward defined in Section~\ref{sec:Method} including penalties and regularization terms. 
whereas the second policy used the complete phase-dependent reward defined in Section~\ref{sec:Method}. The complete reward additionally included the standing-pose, uprightness, and standing-height terms, together with penalties for foot slipping, action magnitude, and action variation.
Tracking performance was evaluated using the aggregate RMS joint-position error over all 29 actuated joints and all inference steps.
Table~\ref{tab:reward_ablation} summarises the aggregate tracking error and task-level performance of both policies for the representative trajectory.
\begin{table}[t]
\centering
\caption{Reward ablation for one representative trajectory}
\label{tab:reward_ablation}
\resizebox{\columnwidth}{!}{
\begin{tabular}{lcc}
\hline
\textbf{Metric} &
\textbf{Tracking only} &
\textbf{Complete reward} \\
\hline
Aggregate RMS error [rad] &
$0.1571$ &
$0.1560$ \\
Final pelvis height [m] &
$0.059$ &
$\mathbf{0.792}$ \\
Maximum pelvis height [m] &
$0.130$ &
$\mathbf{0.794}$ \\
Final uprightness &
$0.543$ &
$\mathbf{0.991}$ \\
Maximum uprightness &
$0.917$ &
$\mathbf{1.000}$ \\
\hline
\end{tabular}}
\end{table}
The two policies achieved nearly identical aggregate joint-position errors:~$0.1571~\mathrm{rad}$ for tracking only and~$0.1560~\mathrm{rad}$ for the complete reward. Nevertheless, their task outcomes differed substantially. The tracking-only policy reached a maximum pelvis height of~$0.130~\mathrm{m}$ and ended at~$0.059~\mathrm{m}$, with uprightness falling from a maximum of~$0.917$ to~$0.543$. In contrast, the complete policy reached and maintained the standing configuration, with a maximum pelvis height of~$0.794~\mathrm{m}$, final height of $0.792~\mathrm{m}$, and final uprightness of~$0.991$.
Hence, accurate joint-space tracking does not guarantee successful whole-body recovery under compliant contact.

Limitations of the proposed approach are as follows. The soft deformable terrain such as foam, grass, or a pile of leaves have compliance in normal as well as tangential directions. Moreover, the robot foot experiences tangential movement resistance to movement because it gets submerged into the terrain. These effects cannot be modelled by MuJoCo's existing contact model. Findings of this study are limited to the simulations.
\section{Conclusions}
This paper presented a reference-guided RL framework for adapting a humanoid stand-up motion from rigid to compliant terrain using a two-stage training procedure.
The method combines residual joint-position control with explicit objectives for pelvis height, uprightness, and final posture under MuJoCo's soft contact dynamics. The learned policies achieved a standing pose on soft as well as hard ground while preserving the overall coordination of their respective reference motions.

The presented simulation results indicate a promising approach for adapting human demos to robot movements on compliant terrain. The future work includes better modelling of terrain compliance and physical validation. The software implementation is made open source to ensure reproducibility and aid further research.
\bibliographystyle{IEEEtran} 
\bibliography{standup_ref}
\end{document}